\documentclass{article}
\usepackage{arxiv}
\usepackage{times}
\usepackage{latexsym}
\usepackage[T1]{fontenc}
\usepackage[utf8]{inputenc}
\usepackage{microtype}
\usepackage{inconsolata}
\usepackage{flushend} 
\usepackage{graphicx}
\usepackage[para,online,flushleft]{threeparttable}
\usepackage{tabularx}
\usepackage{multirow}
\usepackage{booktabs} 
\usepackage{rotating}
 \usepackage{url} 
\usepackage[table]{xcolor}
\usepackage[sort]{natbib}
\usepackage{trimclip}
\usepackage{amsmath}
\usepackage{accents}
\newcommand{\halfapprox}{\clipbox{0em 0em 0em 0.225em}{$\approx$}}

\newcolumntype{s}{>{\columncolor[HTML]{AAACED}} p{3cm}}

\title{MEDEC: A Benchmark for Medical Error Detection and Correction in Clinical Notes}

\author{Asma {Ben Abacha}$^*$\textsuperscript{1}\\
\\\And
  Wen-wai Yim\textsuperscript{1} \\
  \\\And
    Yujuan Fu\textsuperscript{2}\\
 \\\AND
      Zhaoyi Sun\textsuperscript{2}\\
\\\And
   Meliha Yetisgen\textsuperscript{2}\\
  \\\And
   Fei Xia\textsuperscript{2}\\
 \\\And
   Thomas Lin\textsuperscript{1}\\
  \\\AND
\normalfont{\textsuperscript{1} Microsoft, Health and Life Sciences AI, Redmond, USA} 
\\ \textsuperscript{2} University of Washington, Biomedical and Health Informatics, Seattle, USA
\\ \textsuperscript{*} Corresponding author: abenabacha@microsoft.com}

\begin{document}
\maketitle

\begin{abstract}

Several studies showed that Large Language Models (LLMs) can answer medical questions correctly, even outperforming the average human score in some medical exams. However, to our knowledge, no study has been conducted to assess the ability of language models to validate existing or generated medical text for correctness and consistency. In this paper, we introduce MEDEC\footnote{https://github.com/abachaa/MEDEC}, the first publicly available benchmark for medical error detection and correction in clinical notes, covering five types of errors (Diagnosis, Management, Treatment, Pharmacotherapy, and Causal Organism). MEDEC consists of 3,848 clinical texts, including 488 clinical notes from three US hospital systems that were not previously seen by any LLM. The dataset has been used for the MEDIQA-CORR shared task to evaluate seventeen participating systems \citep{mediqa-corr-task}. In this paper, we describe the data creation methods and we evaluate recent LLMs (e.g., o1-preview, GPT-4, Claude 3.5 Sonnet, and Gemini 2.0 Flash) for the tasks of detecting and correcting medical errors requiring both medical knowledge and reasoning capabilities. We also conducted a comparative study where two medical doctors performed the same task on the MEDEC test set. The results showed that MEDEC is a sufficiently challenging benchmark to assess the ability of models to validate existing or generated notes and to correct medical errors. We also found that although recent LLMs have a good performance in error detection and correction, they are still outperformed by medical doctors in these tasks. We discuss the potential factors behind this gap, the insights from our experiments, the limitations of current evaluation metrics, and share potential pointers for future research. 

\end{abstract}
\section{Introduction}

A survey study from US health care organizations showed that 1 in 5 patients who read a clinical note reported finding a mistake and 40\% perceived the mistake as serious, with the most common category of mistakes being related to current or past diagnoses \citep{jama2000}.

\begin{figure*}[h] 
\includegraphics[scale=.47]{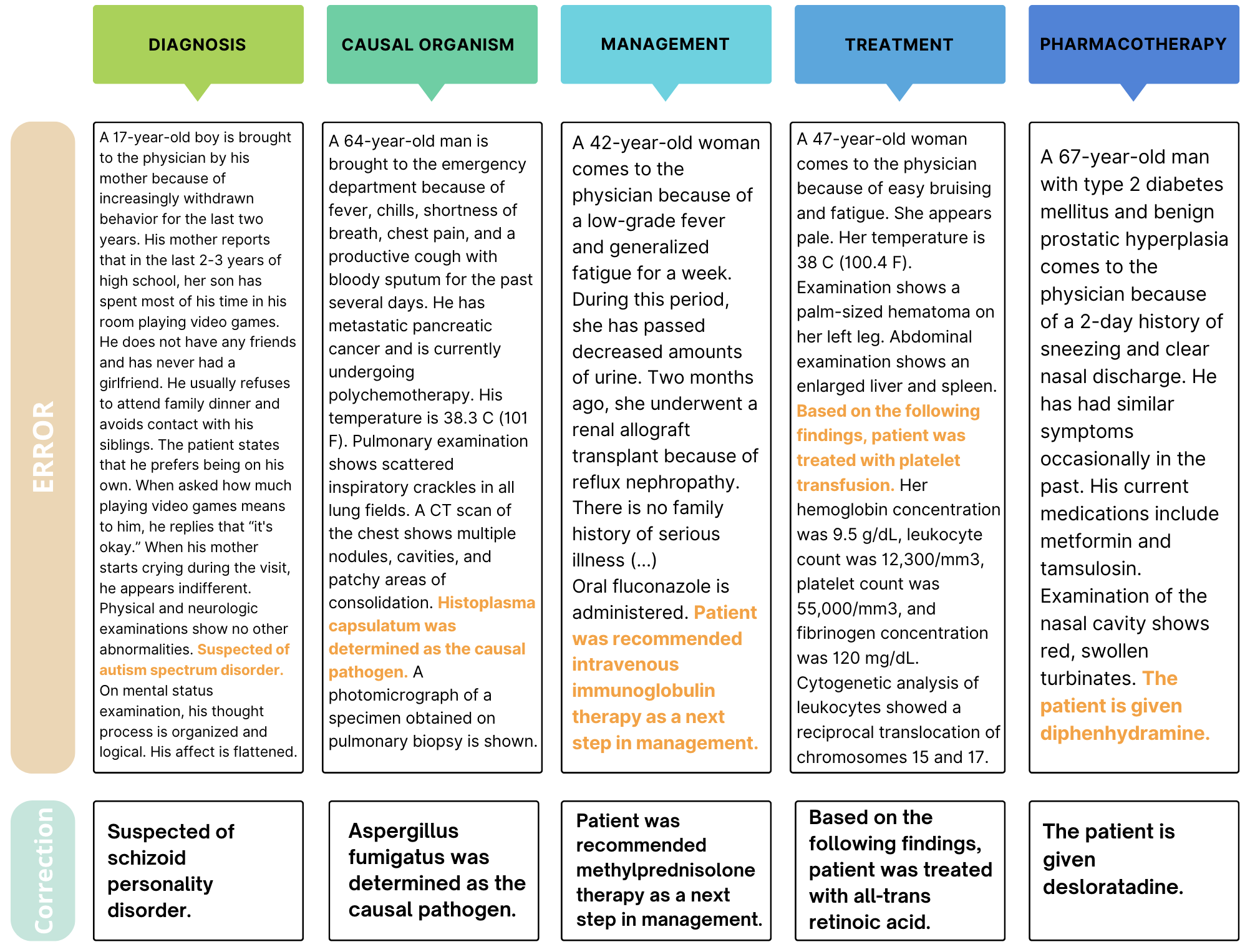}
\caption{Examples from the MEDEC dataset. }
\label{fig:medec}
\end{figure*}

On the other hand, more and more medical documentation tasks (e.g., clinical note generation) are being supported by LLMs. In multiple studies, LLMs have shown the ability to answer accurately questions from medical exams \citep{Gilson2023, Johnson2023, jamanetworkopen2023} and to imitate clinical reasoning in providing diagnoses \citep{Savage0GRC24}. However, one of the main obstacles in adopting LLMs in medical documentation tasks is their potential to generate hallucinations or incorrect information \citep{TangSINSEXDDRWP23} and harmful content that might alter clinical decision making \citep{Chen24}. Rigorous validation methods are essential to mitigate these risks and make LLMs safer to use for medical content generation \citep{Karabacak2023}. 

Such validation requires relevant benchmarks to assess whether it can be fully automated with validation models. A key task in this validation is the ability to detect and correct medical errors in clinical texts. 

Most previous studies on (common sense) error detection have focused on the general domain \citep{wang-etal-2020-semeval,OnoeZCD21}. In this paper, we tackle the problem of identifying and correcting medical errors in clinical texts. From a human perspective, identifying and correcting these errors requires medical expertise, specialized knowledge, and sometimes practical experience. We introduce a new dataset, MEDEC, and experiment with different recent LLMs (e.g., Claude 3.5 Sonnet, o1-preview, and Gemini 2.0 Flash). To the best of our knowledge, this is the first publicly available benchmark and study on automatic error detection and correction in clinical notes. 

\section{Related Work}

\citet{BECEL22} introduced a benchmark for consistency evaluation and evaluated pretrained language models (e.g, BERT, T5, and GPT-2) on three main categories: semantic, logical, and factual consistency. They found that those language models do not perform well in every test case and have a high level of inconsistency in many cases. \citet{JangL23a} investigated the trustworthiness of more recent language models, ChatGPT and GPT-4, regarding semantic consistency and found that while both models appear to show an enhanced language understanding and reasoning ability, they often fail at generating logically consistent predictions. 

In the medical domain, several recent studies evaluated large language models' accuracy and consistency. \citet{Johnson2023} conducted a study to assess the accuracy and reliability of medical responses generated by chatGPT. Thirty-three physicians across 17 specialties generated 284 medical questions with different levels of difficulty and graded ChatGPT's answers for accuracy and completeness. While most of the generated text was evaluated by physicians as accurate, there were potential limitations in handling complex medical questions. 

In two separate studies, \citet{jamanetworkopen2023} and \citet{Gilson2023} found that GPT models can answer medical questions correctly in neurology board–style examinations and the United States Medical Licensing Examination (USMLE) Step 1 and Step 2 exams, even outperforming the average human score in some instances.  

\citet{Chen24} assessed the effect and safety of LLM-assisted patient messaging, as one of the earliest applications of LLMs in electronic health records (EHRs). The increased similarity of response content between LLM drafts and LLM-assisted responses showed that doctors might adopt the LLM's assessment, and that in a minority of LLM drafts, if left unedited, could lead to severe harm or death.

The safe introduction and use of LLMs in medical documentation tasks requires reliable and automatic validation methods. However, as far as we know, no benchmark was made publicly available to assess the ability of LLMs in validating existing or generated medical text for correctness and consistency. 

In this paper, we present MEDEC, the first benchmark for medical error detection and correction in clinical notes. We describe the data creation methods and we evaluate recent state-of-the-art open domain LLMs for these tasks. 

The MEDEC dataset has been used in the first shared task on medical error detection and correction, MEDIQA-CORR 2024, to evaluate models and solutions from  seventeen participating teams \citep{mediqa-corr-task}. 

\section{MEDEC Dataset}

We created a new dataset of 3,848 clinical texts from different specialties. Eight medical annotators participated in the annotation task. The dataset covers five types of errors, selected after analyzing the most frequent types of questions asked in medical board exams. The error types are: 
\begin{itemize}
    \item Diagnosis	- The provided diagnosis is inaccurate.
    \item Management 	-	The next step provided in management is inaccurate.
    \item Pharmacotherapy	-	The recommended pharmacotherapy is inaccurate.
    \item Treatment	-	The recommended treatment is inaccurate.
    \item CausalOrganism	-	The indicated causal organism or causal pathogen is inaccurate.
\end{itemize}

Figure~\ref{fig:medec} presents examples from the MEDEC dataset. Each clinical text is either correct or contains one error created using one of two different methods: M\#1 (MS) \& M\#2 (UW). 

\subsection{Data Creation Method \#1 (MS)}

In this method, we leverage medical board exams from the MedQA collection \citep{MedQA}. Four annotators with medical backgrounds used the medical narratives and multiple choice questions in these exams to inject a wrong answer into the scenario text, after checking the original questions and answers and excluding QA pairs containing errors or ambiguous information. 

\begin{figure*}[h] 
\includegraphics[scale=.47]{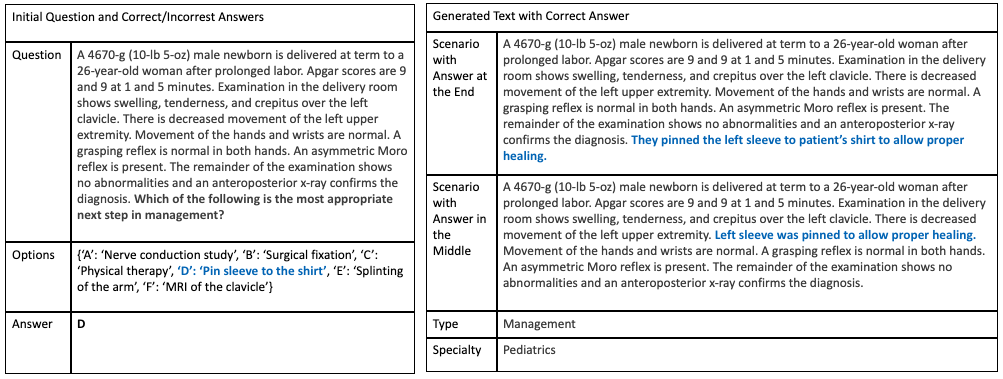}
\caption{Method \#1: Correct answer injected in the question text to create the reference note. The same process was used to inject a selected incorrect answer and to create another version of the note containing a medical error. }
\label{fig:data}
\end{figure*}

The medical annotators followed these guidelines: 
\begin{itemize}
    \item Using medical narrative multiple choice questions, introduce a wrong answer into the scenario text and create two versions with the error injected either in the middle of the text or at the end. 
    \item Using medical narrative multiple choice questions, introduce the right answer into the scenario text to create a correct version, as described in Figure~\ref{fig:data} (\textit{Generated Text with Correct Answer)}. 
    \item Check manually if the automatically rewritten text is faithful to the original scenario and the included answer.
\end{itemize}

We randomly selected one correct and one incorrect version for each note from the two different scenarios (error injected in the middle of the text or at the end) in the final dataset. 

\subsection{Data Creation Method \#2 (UW)}

We used a database of real clinical notes between 2009 and 2021 from three University of Washington (UW) hospital systems\footnote{The MEDEC-UW subset requires signing a DUA. Examples presented in this paper are selected from the MEDEC-MS subset.}: Harborview Medical Center, UW Medical Center, and Seattle Cancer Care Alliance. 

From this database, we randomly selected 488 out of 17,453 diagnosis supports, which summarize patients' medical conditions and provide rationales for treatments. 

A team of four medical students manually introduced errors into 244 of these notes. Initially, each note was marked with several candidate entities identified as Unified Medical Language System (UMLS) \footnote{\url{https://www.nlm.nih.gov/research/umls/licensedcontent/umlsknowledgesources.html}} concepts by QuickUMLS \footnote{\url{https://github.com/Georgetown-IR-Lab/QuickUMLS}}.

An annotator either selected a concise medical entity from these candidates or created a new span. This span was then labeled with one of the five error types. The annotator then replaced this span with an erroneous version using similar but distinct concepts, crafted by the annotators themselves or provided by a SNOMED- and LLM-based method. This method was used to suggest alternative concepts to the annotators without using the input text. Medical annotators decided on the final concepts/errors to inject manually in the text. 

During this process, each error span was required to contradict at least two other parts of the clinical notes (and annotators provided a justification for each error introduced). We de-identified the clinical notes (post error injection) with Philter\footnote{\url{https://github.com/BCHSI/philter-deidstable1_mirror}} for automatic de-identification. Each note was then independently reviewed by two annotators to ensure proper de-identification. A third annotator adjudicated any remaining discrepancies.

The MEDEC dataset contains 3,848 clinical texts. Table~\ref{tab:data-size} presents the training, validation, and test splits. The MS training set contains 2,189 clinical texts. The MS validation set contains 574 clinical texts and the UW validation set contains 160 clinical texts. The MEDEC test set consists of 597 clinical texts from the MS collection and 328 clinical texts from the UW dataset. 51.3\% of the test notes contain errors while 48.7\% of the notes are correct. 

\begin{table}[ht]
\centering
\begin{tabular}{ |lr|c|c|c|c| } 
 \hline
Collection & & Training & Validation & Test & Total \\\hline
MS (M\#1) Subset & \# texts & 2,189 & 574 & 597 & 3,360  \\
UW (M\#2) Subset & \# texts & - & 160 & 328 & 488 \\
\bf MEDEC Dataset & \# texts & \bf 2,189 & \bf 734 & \bf 925 & \bf 3,848 \\
 & \textit{\# texts without errors} & \textit{970 (44.3\%)} &  \textit{335 (45.6\%)} &  \textit{450 (48.7\%)} &  \textit{1,755 (45.6\%)} \\
  & \textit{\# texts with errors} & \textit{1,219 (55.7\%)} &  \textit{399 (54.4\%)} &  \textit{475 (51.3\%)} & \textit{2,093 (54.4\%)} \\\hline
\end{tabular} 
\caption{MEDEC Dataset: Training, Validation, and Test Sets}
\label{tab:data-size}
\end{table}

The MEDEC dataset is available at: \url{https://github.com/abachaa/MEDEC}. The MS subset is publicly available. The UW subset requires signing a data usage agreement (DUA) to have access to the UW notes. \\

\begin{figure*}[h] 
\includegraphics[scale=.68]{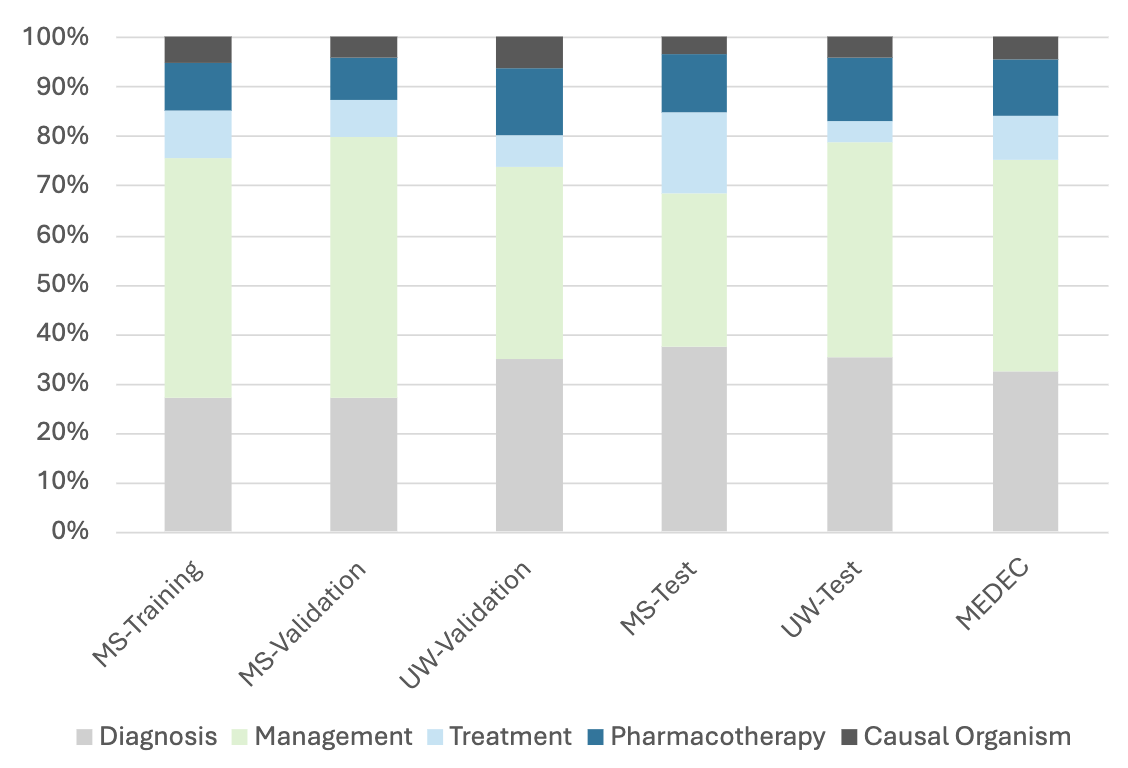}
\caption{Error Type Distribution in the MEDEC Dataset. }
\label{fig:error-types}
\end{figure*}

Figure~\ref{fig:error-types} presents the distribution of error types in the dataset (Diagnosis, Management, Treatment, Pharmacotherapy, and Causal Organism). 

\section{Medical Error Detection \& Correction Approaches}

In order to evaluate models on medical error detection and correction, we divide the process into 3 subtasks: 
\begin{itemize}
    \item \textit{Subtask A:} Predicting the error flag (0: if the text has no error; 1: if the text contains an error). 
    \item \textit{Subtask B:} Extracting the sentence that contains the error for flagged texts (-1: if the text has no error; Sentence ID: if the text contains an error).
    \item \textit{Subtask C:} Generating a corrected sentence for flagged texts with errors (NA: if the text has no error; Generated sentence/correction: if the text has an error). 
\end{itemize}

For comparison, we build LLM-based solutions using two different prompts to generate the outputs required to assess the models on the three subtasks: 
\begin{itemize}
\item \textbf{P\#1}: \textit{\small{The following is a medical narrative about a patient. You are a skilled medical doctor reviewing the clinical text. The text is either correct or contains one error. The text has one sentence per line. Each line starts with the sentence ID, followed by a pipe character then the sentence to check. Check every sentence of the text. If the text is correct return the following output: CORRECT. If the text has a medical error related to treatment, management, cause, or diagnosis, return the sentence id of the sentence containing the error, followed by a space, and then a corrected version of the sentence. Finding and correcting the error requires medical knowledge and reasoning. }}
\item \textbf{P\#2} Similar to the first prompt, but includes an example of input and output, randomly selected from the training set: \textit{\small{Here is an example. 
    0 A 35-year-old woman presents to her physician with a complaint of pain and stiffness in her hands.
    1 She says that the pain began 6 weeks ago a few days after she had gotten over a minor upper respiratory infection. (...)
    9 Bilateral radiographs of the hands demonstrate mild periarticular osteopenia around the left fifth metacarpophalangeal joint.
    10 Methotrexate is given. In this example, the error is in the sentence number 10: Methotrexate is given. The correction is: Prednisone is given. The output is: 10 1 Prednisone is given. End of Example.}}
\end{itemize} 

\section{Experiments \& Results} 

\subsection{Language Models}

We experiment with several recent small and large language models: 
\begin{enumerate} 
    \item Phi-3-7B, a Small Language Model (SLM) with 7 billion parameters \citep{phi3}
    \item Claude 3.5 Sonnet \textit{(2024-10-22)}, the latest model  ($\halfapprox$175B parameters) from the Claude 3.5 family offering state-of-the-art performance across several coding, vision, and reasoning tasks \citep{claude}.
    \item Gemini 2.0 Flash: the latest/most advanced Gemini model \citep{gemini2flash}. Other Google models such as Med-PaLM models (540B) \citep{MedPalm2}, designed for medical purposes, were not publicly available.  
    \item  ChatGPT ($\halfapprox$175B) \citep{chatgpt} and GPT-4 ($\halfapprox$1.76T), a "high-intelligence" model \citep{gpt4}.
    \item GPT-4o ($\halfapprox$200B) providing "GPT-4-level intelligence but faster" \citep{gpt4o} and the GPT-4o-mini \textit{(gpt-4o-2024-05-13)} small model ($\halfapprox$8B parameters) for focused tasks \citep{gpt4o-mini}. 
    \item The latest o1-mini \textit{(o1-mini-2024-09-12)} model ($\halfapprox$100B)  \citep{o1-mini}, and o1-preview \textit{(o1-preview-2024-09-12)} model ($\halfapprox$300B) with "new AI capabilities" for complex reasoning tasks \citep{o1-preview}. 
\end{enumerate} 

The exact numbers of parameters of several LLMs (e.g., GPT, Gemini 2.0 Flash) have not been publicly disclosed yet. 
The model size estimates reported here are mined from public articles only\footnote{\url{https://lifearchitect.ai/o1}, \url{https://www.thealgorithmicbridge.com/p/openai-o1-a-new-paradigm-for-ai}, \url{https://tinyurl.com/mr3c2wme}, \url{https://www.hashtechwave.com/openai-upgrades-explained-o1-preview-o1-mini}, \url{https://felloai.com/2024/08/claude-ai-everything-you-need-to-know}}; the authors cannot vouch for their accuracy and they are provided only to aid in contextualizing model performance. Please refer to the original/future documentation for more precise information about these models.   

Few models (e.g., Phi-3 and Claude) required minimal automatic post-processing to correct some formatting issues. 

\subsection{Evaluation Metrics}

To evaluate the models' performance in recognizing medical errors in texts, we relied on $Accuracy$ for Error Flag Prediction (subtask A) and Error Sentence Detection (subtask B).  

To further analyze the results for each error type, we also computed the $Recall$ using the subset of test examples with errors (error flag = 1) for each type.  

To evaluate the generated corrections (subtask C), we selected lexical and  contextual embedding-based metrics that highly correlate with human judgments on clinical texts \citep{acl2023-eval}. These metrics are: $ROUGE-1$ \citep{rouge}, $BLEURT$ \citep{bleurt}, and $BERTScore$ (microsoft/deberta-xlarge-mnli) \citep{bertscore}. 

We also used an additional score for error correction, Aggregate Score ($AggregateScore$), which is the average of the following evaluation metrics: ROUGE-1, BLEURT-20, and BERTScore.  

We computed these four error correction scores when both the reference and system corrections are provided (other than NA). Our evaluation scripts are available online: \url{https://github.com/abachaa/MEDIQA-CORR-2024/tree/main/evaluation}.

\subsection{Comparison with Expert Labeling} 

Two medical doctors performed the same subtasks on the MEDEC dataset to assess the difficulty of detecting and correcting the errors. The doctors annotated 569 clinical notes from the full test set of 925 texts.  

Given a clinical text from the test set without the ground truth (without the error flag, error sentence, and reference correction), the medical doctors were tasked to: 
\begin{itemize}
\item Judge whether a medical error exists in the text. 
\item If an error exists, write the sentence ID of the sentence where the error occurred. 
\item Provide the most likely error correction and its type (e.g., diagnosis, management, treatment). 
\end{itemize}


\begin{table*}[h!]
  \centering
  \footnotesize
  \setlength{\extrarowheight}{1.0pt}  
  \begin{threeparttable}
    \setlength{\tabcolsep}{2.7mm}{    
        \begin{tabularx}{\textwidth}{l|cc|cccc}
            \toprule 
            \multirow{3}{*}{\textbf{Model}} & \multicolumn{2}{c|}{\textbf{Error Detection Accuracy}} & \multicolumn{4}{c}{\textbf{Error Correction}}   \\ 
            & Error Flag &  Error Sentence & ROUGE-1	& BERTScore	& BLEURT & AggregateScore     \\ \midrule 
             & \multicolumn{6}{c}{\textbf{Small Language Models (7-8B) [P\#1]}} \\ \cmidrule(){2-7}
Phi-3 & 0.5276 & 0.2443 & 0.2606 & 0.1514 & 0.4683 & 0.2935 \\
GPT-4o-mini & 0.6086 & 0.4757 & 0.5148 & 0.5089 & 0.5640 &   0.5292 \\ 
\midrule 
& \multicolumn{6}{c}{\textbf{Large Language Models ($\halfapprox$100-300B)}} \\ \cmidrule(){2-7}
o1-mini & \underline{0.6908} & 0.5968 & 0.6052 & 0.6275 & 0.6246 & 0.6191  \\ %

Claude 3.5 Sonnet & \underline{\underline{0.7016}} & \underline{\underline{0.6562}} &  0.2253 & 0.1033 & 0.5100 & 0.2795   \\ 
Claude 3.5 Sonnet$^{*}$ & 0.6800 & \underline{0.6508} & 0.2249 & 0.1125 & 0.5081 &  0.2818  \\ 
Gemini 2.0 Flash & 0.5805 & 0.3535 &  0.3769 & 0.3127 & 0.4865 & 0.3920  \\ 
ChatGPT & 0.4811 & 0.4800 & 0.4198 & 0.3235 & 0.5133 & 0.4189   \\
GPT-4o & 0.6584 & 0.5665 & 0.5517 & 0.5373 & 0.5852 & 0.4682  \\ 
GPT-4o$^{*}$ & 0.6368 & 0.5449 & 0.5805 & 0.5401 & 0.6022 & 0.5743  \\ 
o1-preview  & 0.6746 & 0.6140 & \underline{\underline{0.6884}} & \underline{\underline{0.7095}} & \underline{\underline{0.6949}} & \underline{\underline{0.6976}}  \\ 
\midrule 
& \multicolumn{6}{c}{\textbf{Large Language Models ($\halfapprox$1.7T)}} \\ \cmidrule(){2-7}

GPT-4 &  0.6573 & 0.5568 & 0.5553 & 0.5804 &  0.5896 & 0.5751   \\
GPT-4$^{*}$ & 0.6519 & 0.5773 & \underline{0.6271} &  \underline{0.6522} & \underline{0.6368} & \underline{0.6387} \\

\bottomrule
& \multicolumn{6}{c}{\textbf{Medical Doctors}} \\ \cmidrule(){2-7}
Doctor \#1 & \bf 0.7961 &  0.6588 & 0.3863 & 0.4653 & 0.5066 & 0.4527   \\  
Doctor \#2&  0.7161 &  \bf 0.6677 & \bf 0.7260 & \bf 0.7315 & \bf 0.6780 & \bf 0.7118   \\ \hline 
        \end{tabularx}
    }
      
        \caption{Accuracy of error (flag \& sentence) prediction and error sentence correction scores. $^{*}$ Uses $P\#2$ prompt. Best LLMs' scores are double underlined. Second best scores are underlined. Best Error Detection Accuracy achieved by Claude  followed by o1-mini (but lower than both doctors' accuracy scores). o1-preview and and GPT-4 [P\#2] achieved the best error correction scores (but lower than Doctor\#2 scores).}
        \label{tab:results-all}
  \end{threeparttable}
\begin{tabular}{l}
\end{tabular}
\end{table*}

\begin{table*}[h!]
  \centering
  \setlength{\extrarowheight}{1.0pt}  
  \footnotesize
  \begin{threeparttable}
    \setlength{\tabcolsep}{2.7mm}{ 
        \begin{tabularx}{\textwidth}{l|cc|cccc}
            \toprule
            \multirow{3}{*}{Dataset}  & \multicolumn{2}{c|}{\textbf{Error Detection Accuracy}} & \multicolumn{4}{c}{\textbf{Error Correction}}  \\ 
            & Error Flag &  Error Sentence &  ROUGE-1	& BERTScore	& BLEURT & AggregateScore   \\            \midrule 

             & \multicolumn{6}{c}{\textbf{Claude 3.5 Sonnet} \textit{(2024-10-22)}} \\ \cmidrule(){2-7}    
MS (M\#1) Subset  & 0.6750 & 0.6348 & 0.1822 & 0.0793 & 0.4996 & 0.2537 \\
UW (M\#2) Subset & \textbf{0.7500} &  \textbf{0.6951} & \textbf{0.3100} & \textbf{0.1508} & \textbf{0.5305} & \textbf{0.3304} \\ 

 \midrule 
             & \multicolumn{6}{c}{\textbf{o1-preview} \textit{(2024-09-12)}} \\ \cmidrule(){2-7}    
MS (M\#1) Subset &  \textbf{0.7286} & \textbf{0.6884} & 0.6857 & \textbf{0.7227} & \textbf{0.7046} & \textbf{0.7043}  \\
UW (M\#2) Subset & 0.5762 & 0.4787 & \textbf{0.6936} & 0.6848 &  0.6767 & 0.6850 \\ 
\bottomrule
& \multicolumn{6}{c}{\textbf{Medical Doctor \#1}} \\ \cmidrule(){2-7} 
MS (M\#1) Subset & \textbf{0.8125} & \textbf{0.7670} & \textbf{0.4199} & \textbf{0.5127} & \textbf{0.5394} & \textbf{0.4907}  \\
UW (M\#2) Subset & 0.7595 & 0.4177 & 0.3073 & 0.3542 & 0.4298 & 0.3638 \\ 
\bottomrule
& \multicolumn{6}{c}{\textbf{Medical Doctor \#2}} \\ \cmidrule(){2-7} 
MS (M\#1) Subset & 0.6890 & 0.6459 & 0.6845 &  0.6981 &  0.6503 & 0.6776  \\
UW (M\#2) Subset & \textbf{0.7723} &  \textbf{0.7129} & \textbf{0.8016} & \textbf{0.7926} & \textbf{0.7284} & \textbf{0.7742}  \\ 
\hline
        \end{tabularx}
    }
    \caption{Accuracy and error correction scores  on each subset: MS \& UW Test Sets. The MEDEC-MS subset was more challenging for Claude and Doctor \#2. The MEDEC-UW subset was more challenging for o1-preview and Doctor \#1.}
    \label{tab:resultsByDataset}
        \vspace{-0.6cm}
  \end{threeparttable}
\end{table*}

\subsection{Results} 
Table~\ref{tab:results-all} presents the results of the manual annotation performed by the medical doctors and the results of several recent LLMs using the two prompts described above. Claude 3.5 Sonnet outperformed the other LLM-based methods in error flag detection with 70.16\% Accuracy and in error sentence detection with 65.62\% Accuracy. 

The o1-mini model achieved the second best error flag detection Accuracy of 69.08\%. In error correction, o1-preview achieved the best Aggregate Score of 0.698, substantially ahead of the second model, GPT-4 [P\#2], with 0.639 Aggregate Score. 

Table~\ref{tab:resultsByDataset} presents the error detection Accuracy and error correction scores on each collection (MEDEC-MS and MEDEC-UW). The MS subset was more challenging for Claude 3.5 Sonnet and Doctor \#2, while the UW subset was more challenging for o1-preview and Doctor \#1. 
 
The results show that recent LLMs have a good performance in error detection and correction, relative to the doctors' scores, but they are still outperformed by the medical doctors in these tasks. This could be explained by the fact that such error detection and correction tasks are relatively rare online and in medical textbooks, which means that these large models are less likely to have encountered such data in their pretraining. This can be seen specifically in the o1-preview results where the model achieved 73\% and 69\% Accuracy in error and sentence detection on the MS subset that was built from publicly available clinical texts, while achieving only 58\% and 48\% Accuracy on the private UW collection. 

Another factor is that the task consists in analyzing and fixing an existing text that was not generated by LLMs, which might have a higher level of difficulty than drafting new answers from scratch.  

Table~\ref{tab:resultsByType}  presents the error detection Recall and error correction scores for each error type (Diagnosis, Management, Treatment, Pharmacotherapy, and Causal Organism). The o1-preview model had substantially higher error flag and sentence detection Recall scores across all error types compared to Claude 3.5 Sonnet and both doctors. Combined with the overall Accuracy results (cf. Table~\ref{tab:results-all}), where the doctors achieved better Accuracy, these results indicate that the model(s) had a substantial issue on the Precision side and hallucinated error presence in many cases compared to medical doctors.  

The results also show that there is a ranking discrepancy between classification performance and error correction generation performance. For instance, Claude 3.5 Sonnet was first in Accuracy of error flag and sentence detection among all the models, but was last in correction generation scores (cf. Table~\ref{tab:results-all}). Also, o1-preview was fourth in error detection Accuracy among all the LLMs, but was first and substantially ahead in correction generation. The same pattern could be observed between the two medical doctors. 

\begin{table*}[h!]
  \centering
  \setlength{\extrarowheight}{1.0pt}  
  \footnotesize
  \begin{threeparttable}
    \setlength{\tabcolsep}{2.7mm}{ 
        \begin{tabularx}{\textwidth}{l|cc|cccc}
            \toprule
            \multirow{3}{*}{\textbf{Error Type}} & \multicolumn{2}{c|}{\textbf{Error Detection Recall}} & \multicolumn{4}{c}{\textbf{Error Correction}}   \\
            & Error Flag &  Error Sentence &  ROUGE-1	& BERTScore	& BLEURT & AggregateScore    \\ 
            \midrule 
             & \multicolumn{6}{c}{\textbf{Claude 3.5 Sonnet} \textit{(2024-10-22)}} \\ \cmidrule(){2-7} 
Diagnosis & 0.5977 & 0.5344 &  0.2416 &  0.1051 & 0.5390 & 0.2953 \\
Management  &  0.6131 & 0.4881 & 0.2157 & 0.0968 & 0.4877 & 0.2667  
\\
Treatment & 0.6034 & 0.5345 &  0.1607 & 0.0831 & 0.4890 & 0.2442  \\
Pharmacotherapy & 0.7017 & 0.6316 & 0.2577 & 0.1401 & 0.5089 & 0.3023  \\
Causal Organism & 0.8333 & 0.7222 & 0.2422 & 0.0851 & 0.5130 & 0.2801  \\ 
\bottomrule 

& \multicolumn{6}{c}{\textbf{o1-preview} \textit{(2024-09-12)}} \\ \cmidrule(){2-7} 
Diagnosis & \textbf{0.9655} & \textbf{0.8391} & 0.7706 & 0.7852 & \textbf{0.7447} &  0.7668 \\
Management  & \textbf{0.9345} & \textbf{0.7679} & 0.5697 &  0.6039 &  0.6125 & 0.5954  \\
Treatment & \textbf{0.9310} & \textbf{0.8965} &  \textbf{0.7034} & \textbf{0.7628} & \textbf{0.7207} & \textbf{0.7289}  \\
Pharmacotherapy &  \textbf{0.9649} & \textbf{0.8947} & 0.7536 & 0.7369 & \textbf{0.7406} & \textbf{0.7437}  \\
Causal Organism & \textbf{1.0} & \textbf{1.0} & \textbf{0.7131} & \textbf{0.6802} & \textbf{0.7318} & \textbf{0.7084}  \\ 

\bottomrule
& \multicolumn{6}{c}{\textbf{Medical Doctor \#1}} \\ \cmidrule(){2-7} 
Diagnosis & 0.8333 & 0.6863 & 0.4810 & 0.5616 & 0.5668 & 0.5365 \\  
Management & 0.8267 & 0.6000 & 0.2788 &  0.3375 & 0.4371 & 0.3511  \\  
Treatment & 0.7200 & 0.6800 & 0.2726 & 0.4032 & 0.4316 & 0.3691  \\  
Pharmacotherapy & 0.8000 & 0.7200 & 0.4377 & 0.5319 &  0.5371 & 0.5022    \\  
Causal Organism & 0.7273 & 0.7273 &  0.3664 & 0.4309 & 0.5090 & 0.4354  \\  
\bottomrule
& \multicolumn{6}{c}{\textbf{Medical Doctor \#2}} \\ \cmidrule(){2-7} 
Diagnosis & 0.7232 & 0.6786 & \textbf{0.8121} & \textbf{0.8128} & 0.7413 & \textbf{0.7887}  \\  
Management & 0.6893 & 0.6311 & \textbf{0.6763} & \textbf{0.6774} &  \textbf{0.6487} & \textbf{0.6675}  \\  
Treatment & 0.7273 & 0.6970 & 0.5594 & 0.6147 & 0.5770 & 0.5837   \\ 

Pharmacotherapy & 0.8182 & 0.7576 & \textbf{0.7592} & \textbf{0.7464} & 0.6774 & 0.7277  \\  
Causal Organism & 0.4286 & 0.2857 & 0.4474 & 0.4632 & 0.4141 & 0.4415  \\  
\hline 
        \end{tabularx}
    }
    \caption{Recall and error correction scores for each error type using the subset of test examples with errors. The size of each reference subset is as follows: Diagnosis (174 texts), Management (168), Treatment (58), Pharmacotherapy (57), and Causal Organism (18).}
    \label{tab:resultsByType}
    \vspace{-0.7cm}
  \end{threeparttable}
\end{table*}


Part of it could be explained by the difficulty of the correction generation task, but also, the limitations of current SOTA text generation evaluation metrics in capturing synonyms and similarities in medical texts.  

Table~\ref{tab:examples} presents examples from the reference texts, doctors' annotations, and automatically generated corrections by Claude 3.5 Sonnet and GPT models. For instance, the reference correction of the second example indicates that the patient is diagnosed with \textit{Bruton agammaglobulinemia}, while the LLMs provided correct answers mentioning \textit{X-linked agammaglobulinemia} (a synonym of the same rare genetic disease). Also, some LLMs such as Claude provide long answers/corrections with more explanation. Similar observations can be found within the doctors' annotations, where Doctor \#1 provided longer corrections than Doctor \#2, and both doctors had different opinions in some examples/cases, reflecting some of the differences in style and content found in clinical notes written by different doctors/specialists.

Our observations suggest that relevant research efforts on medical error detection and correction might include using more examples and example adaptation in the prompts. In this introductory work, we focused on evaluating state-of-the-art open domain LLMs, but further investigations should also include specialized medical language models as well as explore new evaluation metrics for clinical texts \citep{acl2023-eval}.  

\begin{table*}[ht!]
\centering
\scriptsize
\begin{tabular}{l|c|c||c|c||c|c}
\hline
& \multicolumn{2}{c||}{\textbf{MEDEC Dataset (Text with Error and its Correction)}} &  \multicolumn{2}{c||}{\textbf{Doctors' corrections}} &  \multicolumn{2}{c}{\textbf{LLMs' corrections}} \\
\hline
& \bf Text with Error highlighted in Bold & \bf Reference & \bf Doctor \#1 & \bf Doctor \#2  & \bf GPT & \bf Claude\\
\hline

\parbox[t]{0.001in}{\begin{turn}{-90}Diagnosis (ms-test-141) \end{turn}} & \parbox[t]{1.4in}{A 23-year-old man presents with sudden loss of consciousness while pitching in a baseball game. There is no history of injury. Consciousness is regained after cardiopulmonary resuscitation. Past medical history is negative for any neurological and cardiovascular problems. \textbf{Patient is diagnosed with \colorbox{pink}{aortic stenosis} after physical examination reveals a prominent A wave on the jugular venous pulse and a double apical impulse.} There are no audible murmurs. An S4 is present.}
         & \parbox[t]{0.7in}{Patient is diagnosed with \colorbox{green}{hypertrophic} \colorbox{green}{cardiomyopathy} after physical examination reveals a prominent A wave on the jugular venous pulse and a double apical impulse.}  & \parbox[t]{0.6in}{The patient s diagnosed with \underline{hypertrophic} \underline{cardiomyopathy} afte physical examination reveal a double apical impulse, prominent "a wave" and S4.}  & \parbox[t]{0.7in}{Patient is diagnosed with \underline{hypertrophic} \underline{cardiomyopathy} after physical examination reveals a prominent A wave on the jugular venous pulse and a double apical impulse.}  & \parbox[t]{0.7in}{ \textit{[GPT-4 \& GPT-4o \& o1-preview \& o1-mini]} Patient is diagnosed with \underline{hypertrophic} \underline{cardiomyopathy} after physical examination reveals a prominent A wave on the jugular venous pulse and a double apical impulse.} & \parbox[t]{0.9in}{\cellcolor[HTML]{C0C0C0} \tiny{A systolic ejection murmur is present at the right upper sternal border.  The error is in sentence 5. In aortic stenosis, a characteristic systolic ejection murmur is typically heard best at the right upper sternal border (2nd right intercostal space) and often radiates to the carotid arteries. This is one of the key diagnostic features of aortic stenosis, along with the other findings mentioned in the case (prominent A wave in JVP, double apical impulse, and S4)...}} \\ \hline %

\parbox[t]{0.001in}{\begin{turn}{-90}Diagnosis (ms-test-234)\end{turn}} & \parbox[t]{1.4in}{A 4-year-old boy presents with a history of recurrent bacterial infections, including several episodes of pneumococcal sepsis. His 2 maternal uncles died after having had similar complaints. \textbf{Patient is diagnosed with \colorbox{pink}{common variable immunodeficiency} after lab investigations reveal an undetectable level of all serum immunoglobulins.}}  & \parbox[t]{0.8in}{Patient is diagnosed with \colorbox{green}{Bruton} \colorbox{green}{agammaglobulinemia} after lab investigations reveal an undetectable level of all serum immunoglobulins.}  & \parbox[t]{0.6in}{\cellcolor[HTML]{C0C0C0} The differential diagnosis includes  autosomal recessive agammaglobulinemia, but further evaluation is necessary.}  & \parbox[t]{0.8in}{Patient is diagnosed with \underline{X-linked} \underline{agammaglobulinemia} after lab investigations reveal an undetectable level of all serum immunoglobulins.} & \parbox[t]{0.8in}{\textit{[GPT-4 \& GPT-4o \& o1-preview \& o1-mini]} Patient is diagnosed with \underline{X-linked} \underline{agammaglobulinemia} after lab investigations reveal an undetectable level of all serum immunoglobulins.  } & \parbox[t]{0.9in}{\tiny{Patient is diagnosed with \underline{X-linked} \underline{agammaglobulinemia} after lab investigations reveal an undetectable level of all serum immunoglobulins. The error is in sentence 2. Given the clinical presentation (young male patient with recurrent bacterial infections), family history (maternal uncles affected, suggesting X-linked inheritance), and laboratory findings...}} \\ \hline 

\parbox[t]{0.001in}{\begin{turn}{-90}Management (ms-test-261)\end{turn}} & \parbox[t]{1.4in}{A 50-year-old woman visits her primary care practitioner with the complaints of generalized weakness, lightheadedness, and fatigability for the past month. She also claims to have epigastric pain, heartburn, and a sensation of a lump in her throat. Her family history is irrelevant, but her medical history is significant for a Pap smear that reported atypical squamous cells of undetermined significance (ASCUS), which was followed up with a cervical biopsy negative for malignancy. She occasionally takes over-the-counter medicines to cope with the heartburn, and recently her hemoglobin was found to be 11 g/dL, for which she received iron and vitamin B12 supplementation. \textbf{\colorbox{pink}{Fecal occult blood test} is completed.} Physical examination is unremarkable, except for pale skin, and a pulse of 120/min.}  & \parbox[t]{0.7in}{The patient is referred for an \colorbox{green}{endoscopy}.}  & \parbox[t]{0.6in}{She has not had followup after her cervical biopsy and HPV testing is indicated.}  & \parbox[t]{0.8in}{ \cellcolor[HTML]{C0C0C0} Text annotated as CORRECT} & \parbox[t]{0.8in}{ \cellcolor[HTML]{C0C0C0} \textit{[GPT-4-P1]} Physical examination is unremarkable, except for pale skin, and a pulse of 100/min.  \textit{[GPT-4-P2 \& GPT-4o-P1 \& GPT-4o-P2]} Text annotated as CORRECT.  \textit{[GPT-4o-mini \& o1-preview \& o1-mini]} She occasionally takes over-the-counter medicines to cope with the heartburn, and recently her hemoglobin was found to be 11 g/dL, for which she received iron supplementation. } & \parbox[t]{0.9in}{ \cellcolor[HTML]{C0C0C0}Text annotated as CORRECT} \\ \hline 

\end{tabular}
\caption{Examples of manual \& automatic corrections. Incorrect annotations/outputs are highlighted in Grey.}
\label{tab:examples}
\end{table*}



\section{Conclusion}

This paper presented the MEDEC benchmark for medical error detection and correction in clinical notes. An empirical evaluation of LLM-based methods showed that, while recent LLMs have a good performance, they are still outperformed by medical doctors. The results of the doctors' annotation showed that the MEDEC dataset is a sufficiently challenging benchmark to assess the ability of models to validate existing or generated notes and to correct medical errors. We hope that this dataset will enable further studies on medical error detection and correction in clinical notes, enhancing clinical reasoning capabilities of LLMs, and facilitate additional efforts on evaluation metrics for clinical texts and applications. 

\section{Limitations}

The paper does not cover all types of possible methods and models for the detection and correction of medical errors. The dataset is also limited in terms of size and types of medical errors. 

\section*{Acknowledgements} 
We thank the doctors who participated in this study as well as our annotation team (Erica Labrie, Loren Kimmel, Seanjeet Paul, Thomas Ryan, Brianna L Cowin, Sabrina J Crooks, Karina Lopez, and Kelsi F Nabity).

\bibliography{references}

\end{document}